\documentclass{myownws-acs}
\usepackage{url,bm,color,graphicx}

\begin{document}

\newcommand{\beq}{\begin{equation}}
\newcommand{\eeq}{\end{equation}}
\newcommand{\abs}[1]{\vert#1\vert}
\newcommand{\e}{{\rm e}}
\newcommand{\eps}{\varepsilon}
\newcommand{\mean}[1]{{\langle{#1}\rangle}}
\newcommand{\s}{\sigma}
\renewcommand{\t}{\tau}
\newcommand{\w}[1]{{\overline{#1}}}
\newcommand{\ant}{{\rm ant}}
\newcommand{\dy}{{\rm dyn}}
\newcommand{\st}{{\rm st}}
\renewcommand{\th}{{\rm th}}
\newcommand{\prob}[1]{{\rm Prob}\{#1\}}


\markboth{Mehta and Luck}{Hearings and mishearings: decrypting the spoken word}

%
%

\title{HEARINGS AND MISHEARINGS:\\ DECRYPTING THE SPOKEN WORD}

\author{ANITA MEHTA}

\address{Centre for Linguistics and Philology, University of Oxford,\\
Walton Street, Oxford OX1 2HG, UK\\
anita.mehta@ling-phil.ox.ac.uk}

\author{JEAN-MARC LUCK}

\address{Universit\'e Paris-Saclay, CNRS, CEA, Institut de Physique Th\'eorique,\\
91191 Gif-sur-Yvette, France\\
jean-marc.luck@ipht.fr}

\maketitle

\begin{history}
\end{history}

\begin{abstract}
We propose a model of the speech perception of individual words in the presence
of mishearings.
This phenomenological approach is based on concepts used in linguistics,
and provides a formalism that is universal across languages.
We put forward an efficient two-parameter form for the word length distribution,
and introduce a simple representation of mishearings, which we use in our
subsequent modelling of word recognition.
In a context-free scenario, word recognition often occurs via anticipation when,
part-way into a word, we can correctly guess its full form.
We give a quantitative estimate of this anticipation threshold when no
mishearings occur, in terms of model parameters.
As might be expected, the whole anticipation effect disappears when there are
sufficiently many mishearings.
Our global approach to the problem of speech perception is in the spirit of an
optimisation problem.
We show for instance that speech perception is easy when the word length is
less than a threshold,
to be identified with a static transition, and hard otherwise.
We extend this to the dynamics of word recognition,
proposing an intuitive approach
highlighting the distinction between individual, isolated mishearings and
clusters of contiguous mishearings.
At least in some parameter range, a dynamical transition is manifest well
before the static transition is reached,
as is the case for many other examples of complex systems.
\end{abstract}


\section{Introduction}

Any language is at once the vehicle of poetry and literature,
and a precise algorithm for communication between its speakers,
whose efficacy depends, among other things, on the size of its lexicon and the
complexity of its grammar.
Any aspect of language, be it its intrinsic structure or its decryption,
is then as much a matter of science as of art:
in this sense, the study of languages is one of the first instances of truly
interdisciplinary academic activity.
The analysis of languages accordingly includes approaches that range from being
very descriptive
and instance-specific, to ones that are global and quantitative.
An early instance of the latter concerns studies of the word length
distribution in various texts and languages,
whose historical roots date back at least to the middle of the 19th century
(see~Chapter~2 of~\cite{grzybek} for a comprehensive review).
Nowadays extensive databases such as the Leipzig Corpora
Collection~\cite{leipzig} are available,
allowing a comparison between models and~data.

The aim of this work is to model
the alteration of spoken words by mishearings and the decrypting of speech in
the presence of mishearings.
Our approach is that of statistical physics.
Earlier applications of concepts and tools from statistical physics to various
aspects of linguistics
are reviewed in~\cite{fcs,kwapien,blythe,altmann}.
We endeavour to make a minimal model, keeping only the essential features of the problem.
Among the most important linguistic concepts that we draw upon is the notion of
`underspecification'~\cite{lahiri1} in speech, which increases the efficiency
of the decryption processes;
this is done by incorporating an intermediate stage where mishearings are
allowed for in the construction of a possible word.

The plan of this paper is as follows.
Section~\ref{model} contains some preliminary material to be used in subsequent
developments.
We present our modeling of mishearings and the ensuing statistics of word
variants (Section~\ref{variants}),
as well as a very efficient two-parameter representation of the word length
distribution across languages (Section~\ref{wordlength}).
Our main results on speech decryption are presented in the two following sections.
Section~\ref{wordrec} is devoted to a phenomenological analysis of the statics
of word recognition,
including the highlighting of a static easy-to-hard transition
(Section~\ref{e2h})
and the analysis of the anticipation effect in the absence and in the presence
of mishearings (Sections~\ref{anti0} and~\ref{anti1}).
A more intuitive investigation of dynamical aspects of word recognition is
presented in Section~\ref{dynamic}.
Section~\ref{discussion} contains a brief discussion of our findings.
An appendix is devoted to the statistical mechanics of chains with
intra-cluster interactions.

\section{Spoken words and their variants}
\label{model}

In the branches of linguistics devoted to spoken language, i.e., phonetics and phonology
(see~\cite{grzybek,hay,lad,G+J} for overviews),
the smallest distinctive unit of speech is called a phoneme.
These phonemes, divided into vowels and consonants, are specific to a given
language and range between a minimum
of 11 and a maximum of 160~\cite{hay} across world languages.
Typically, languages have 20 to 40 phonemes; English, for example, has 44.

Lahiri and co-workers~\cite{lahiri2} sought to make this deconstruction more universal,
by relating perceived sounds to anatomical {\it features}, so that
language-independent frameworks for analysis could be set up.
The universality of this scheme provided the inspiration for our model below,
which is also set up in terms of a language-independent formalism.
That said, this first attempt is far from incorporating the subtle details of features.
We refer instead to the elementary units of speech simply as `sounds' which are
meant to be universal across languages,
along the lines suggested by Lahiri et al~\cite{lahiri2}.
Another core assumption in our model is that the linguistic lexicon is {\it structureless},
in the sense that the different sounds that form a word are not correlated
among themselves.
This provides a useful simplified framework for our investigations.

\subsection{Mishearings and word variants}
\label{variants}

This section is devoted to our modeling of mishearings.
For that purpose it is sufficient to deal with the simplest of all possible lexicons,
where a word of length~$n$,
\beq
w=\s_1\dots\s_n,
\label{wdef}
\eeq
is nothing but an arbitrary sequence of $n$ sounds,
where each sound $\s_i$ ($i=1,\dots,n$) is chosen among the $\nu$ sounds of the
language under consideration.
Within this framework, the number $W_n$ of `unique' (i.e., distinct) words of length $n$,
\beq
W_n=\nu^n,
\label{wordnb}
\eeq
grows exponentially with the word length.
This oversimplified situation giving rise to an exponential proliferation of
words will be replaced, from Section~\ref{wordlength} onward,
by the more realistic setting of a finite lexicon and a non-trivial,
language-specific word length distribution.

We model mishearings as independent, random and local alterations of sounds.
In the presence of mishearings, the spoken word $w$ is thus heard as a `variant'
\beq
\widetilde w=\t_1\dots\t_n,
\eeq
where each sound $\s_i$ in the word $w$ can either be misheard (with probability $q$)
or correctly heard (with probability $1-q$).
The mishearing probability $q$ is one of the key parameters of this work.
We further assume for simplicity that each sound can only be misheard in one
particular way.
For instance, the sound `n', if misheard, will always be perceived as `m'.
These rules may be summarised as
\beq
\t_i=\left\{
\begin{matrix}
\phi(\s_i)\quad & \mbox{with prob.}\;q,\hfill\cr
\s_i\hfill & \mbox{with prob.}\;1-q.\cr
\end{matrix}
\right.
\label{rules}
\eeq
The alteration function $\phi$ encodes the allowed mishearing of each sound;
for instance, $\phi(\mbox{`n'})=\mbox{`m'}$.

For a word $w$ of length $n$, i.e., consisting of $n$ sounds,
if exactly $k$ sounds at specified positions are misheard, the
rules~(\ref{rules}) generate
\beq
\omega=2^k
\eeq
distinct word variants $\widetilde w$.
For instance, for the word $w=\s_1\s_2\s_3$ of length $n=3$,
if the last two sounds are misheard, we have $k=2$ and $\omega=4$.
The four variants read
\beq
\widetilde w=\s_1\s_2\s_3,\quad
\widetilde w=\s_1\phi(\s_2)\s_3,\quad
\widetilde w=\s_1\s_2\phi(\s_3),\quad
\widetilde w=\s_1\phi(\s_2)\phi(\s_3).
\eeq

The distribution of the number $\omega$ of variants of a given word $w$ of length $n$
ensues from the observation that the number $k$ of mishearings is distributed
according to the binomial law:
\beq
\prob{k}={n\choose k}q^k(1-q)^{n-k}\quad(k=0,\dots,n).
\label{binom}
\eeq
One key quantity in subsequent developments is the average number of variants
of a word of length $n$:
\beq
\Omega_n=\mean{\omega}=\sum_k2^k\,\prob{k}=\e^{\kappa n},
\label{omegaave}
\eeq
with
\beq
\kappa=\ln(1+q).
\label{kappa}
\eeq
The average number of variants per word thus also grows exponentially with the
word length.
The relationship between the numbers of words (see~(\ref{wordnb})) and their
variants (see~(\ref{omegaave}))
can be usefully expressed in terms of a scaling exponent $\delta$, such that
\beq
\Omega_n\sim W_n^\delta,
\eeq
i.e.,
\beq
\delta=\frac{\kappa}{\ln\nu}.
\label{delta}
\eeq
For typical parameter values such as $q=0.2$ and $\nu=20$, we obtain
\beq
\delta\approx0.06.
\label{deltazero}
\eeq
The smallness of this value suggests that the number of variants is not
overwhelmingly large,
so that even with mishearings words are typically rather easy to recognise.

As an alternative to~(\ref{omegaave}), one could instead consider the most
probable number of variants of a word of length $n$,
defined according to the usage in the statistical physics of disordered
systems~\cite{by,mpv}:
\beq
\overline\Omega_n=\e^{\mean{\ln\omega}}=\e^{\overline\kappa n},
\label{omegamax}
\eeq
with
\beq
\overline\kappa=q\ln 2.
\label{kappabar}
\eeq
More generally, all the moments of the number $\omega$ of variants per word
grow exponentially with the word length.
We have indeed
\beq
\mean{\omega^s}=\sum_k2^{sk}\,\prob{k}=\e^{\lambda(s)n},
\label{omegas}
\eeq
where the exponent
\beq
\lambda(s)=\ln(1+(2^s-1)q)
\label{kappaofs}
\eeq
depends non-linearly on the order $s$ of the moment under consideration.
Here $s$ is not necessarily an integer.
We have consistently $\kappa=\lambda(1)$ and $\overline\kappa=\lambda'(0)$,
where the prime denotes a derivative.
A scaling law such as~(\ref{omegas}), with a non-linear dependence of the
exponent $\lambda(s)$,
is reminiscent of multifractal analysis~\cite{paladin,stanley}.

In the present situation, $\lambda(s)\approx(2^s-1)q$ vanishes linearly with
the mishearing probability as $q\to0$, irrespective of the order $s$.
This justifies the use of the average number $\Omega_n$ of word variants in the
subsequent analysis of various quantities.
Considering another moment of that number, including the most probable one ($s\to0$),
would indeed essentially amount to rescaling the mishearing probability $q$ by
a constant factor.

\subsection{Word length distributions}
\label{wordlength}

In this section we introduce an efficient parametrisation of the word length
distribution across natural languages,
which will be used in what follows to refine the calculations presented in
Section~\ref{variants} of word variants generated by mishearings.

\begin{figure}[!ht]
\begin{center}
\includegraphics[angle=0,width=.55\linewidth,clip=true]{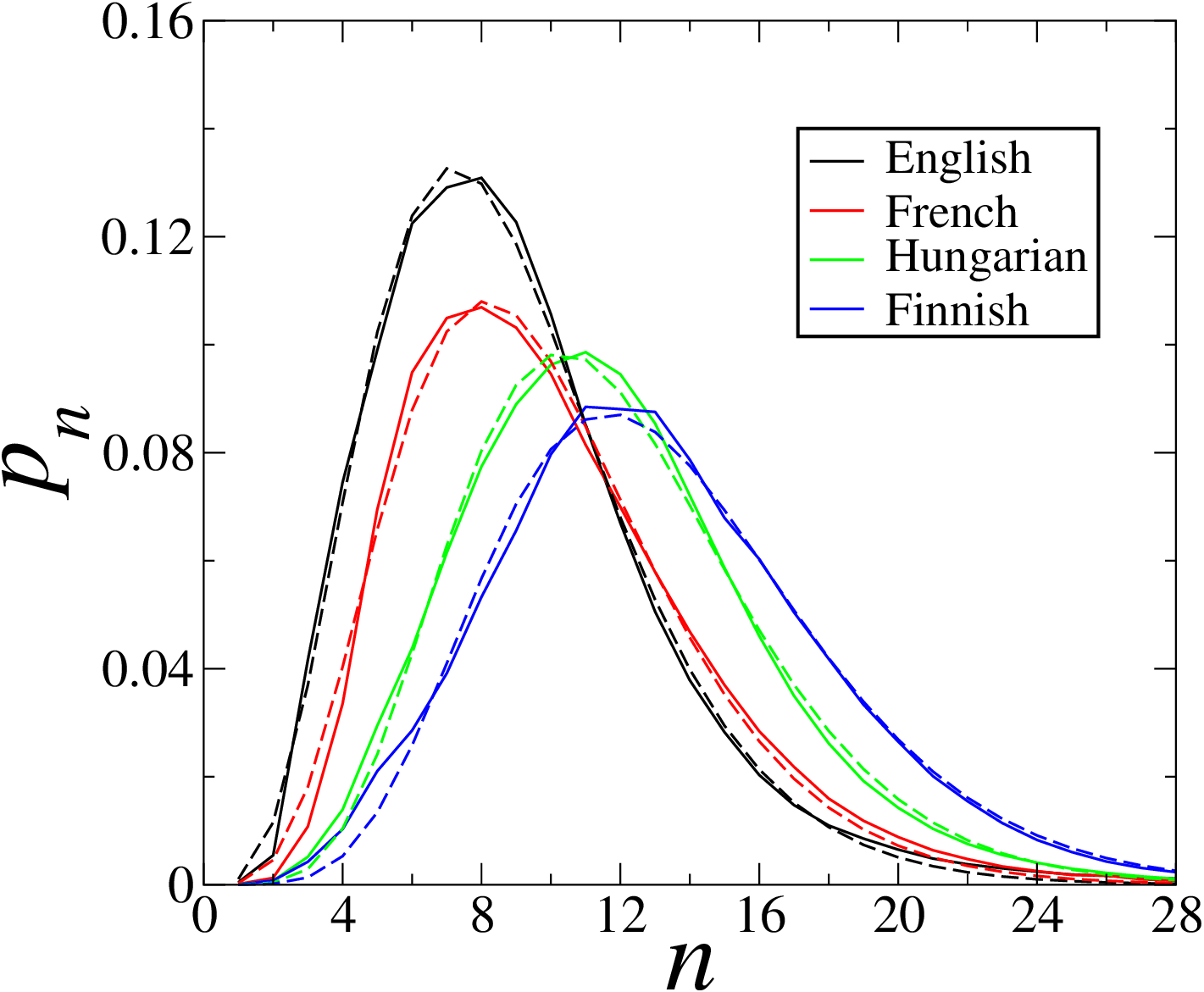}
\vskip 6pt
\includegraphics[angle=0,width=.55\linewidth,clip=true]{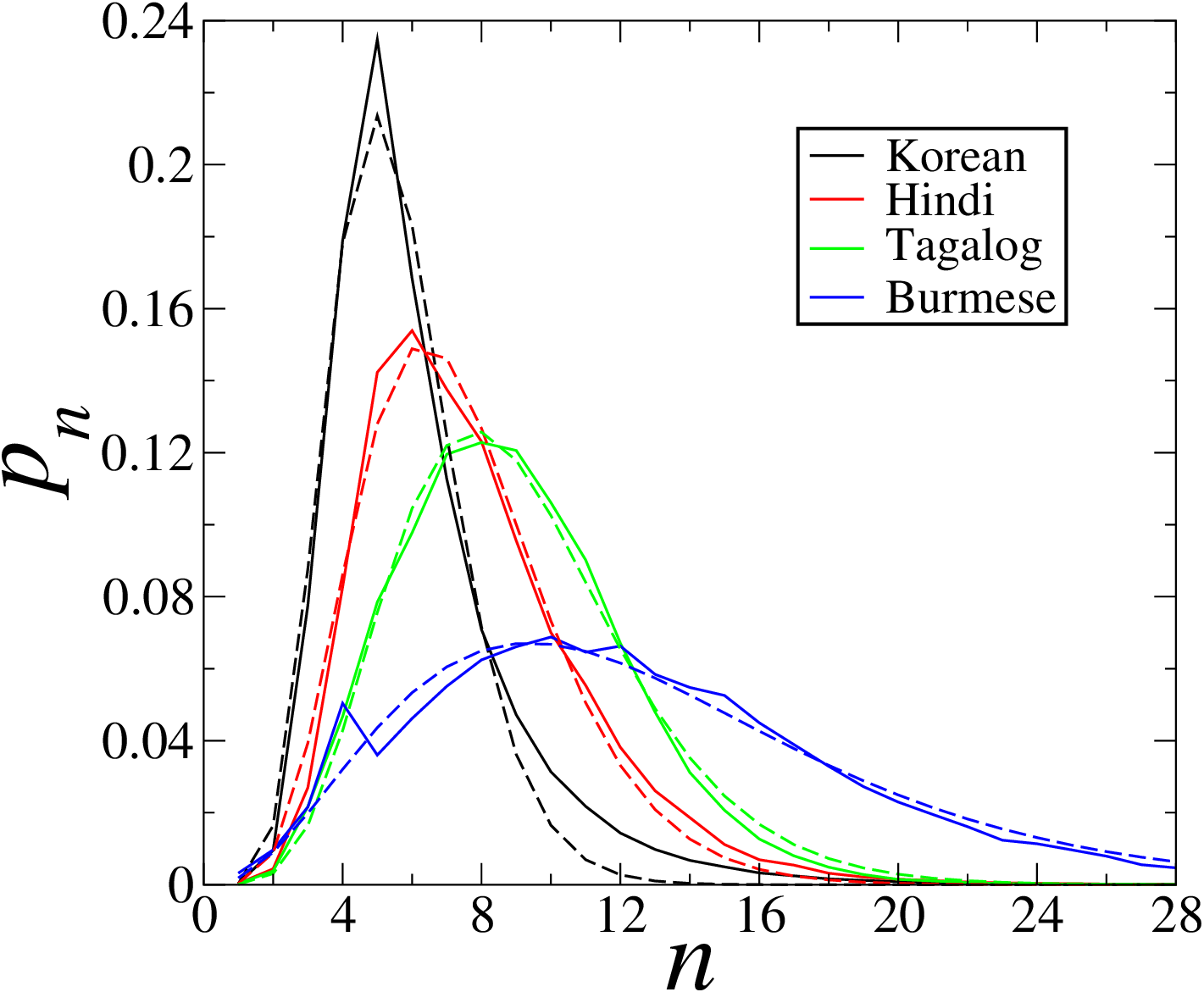}
\caption{
Distribution of the length of written words for four European languages (upper panel)
and four Asian languages (lower panel).
Full curves: data from the Leipzig Corpora Collection~\cite{leipzig}.
Dashed curves: two-parameter fits to~(\ref{gamma}).}
\label{wldplot}
\end{center}
\end{figure}

The study of the distribution of word lengths is an old subject
(see~Chapter~2 of~\cite{grzybek} for a comprehensive review).
An overwhelming majority of the available data concerns the length distribution
of unique written words,
where each word of the lexicon is counted once,
and its length is defined as the number of its letters.
A broad variety of parametrisations of the latter distribution has been proposed,
ranging from the simplest (geometric, Poissonian, log-normal)
to arbitrarily complex multi-parameter distributions.
It appears that word length distributions in most current languages are
characterised by two main features:
a very rapid initial increase and a more or less steep decay with size.
In this work we put forward the use of the Gamma distribution
\beq
p_n=C\,n^\alpha\,\e^{-\beta n}
\label{gamma}
\eeq
as an efficient way to describe these features.
This distribution has a minimal number of two parameters, $\alpha$ and $\beta$,
describing respectively the rise and the fall-off of the distribution.
In spite of this simplicity,
the Gamma distribution~(\ref{gamma}) has hardly ever been used to parametrise
word length distributions in their generality (see~\cite{gamma1,gamma2,gamma3}
for its use in specific instances).
In Figure~\ref{wldplot}, we demonstrate that it provides a very good
parametrisation of the full word length distribution
for four European languages (upper panel) and four Asian languages (lower panel),
which have between them a broad range of average word lengths.
The dashed curves show fits to~(\ref{gamma}), to be compared with empirical
data (full curves)
for the length distributions of unique written words provided by the Leipzig
Corpora Collection~\cite{leipzig}.
The corresponding fitted parameter values $\alpha$ and $\beta$
are given in Table~\ref{one}.
Fits of similar quality have been obtained across many other languages which we
have considered.
We therefore argue that the Gamma distribution~(\ref{gamma}) has a universal
validity across most world languages.

\begin{table}[!ht]
\begin{center}
\begin{tabular}{|l|c|c|c|c|}
\hline
Language & $\alpha$ & $\beta$ & $\w n$ & $n_*$ \\
\hline
English & 4.4 & 0.60 & 8.3 & 7 \\
French & 4.9 & 0.60 & 10.1 & 8 \\
Hungarian & 6.7 & 0.64 & 11.9 & 10 \\
Finnish & 6.8 & 0.58 & 13.4 & 12 \\
\hline
Korean & 8.2 & 1.65 & 6.2 & 5 \\
Hindi & 6.0 & 0.94 & 7.7 & 6 \\
Tagalog & 6.0 & 0.73 & 8.8 & 8 \\
Burmese & 2.6 & 0.28 & 12.8 & 9 \\
\hline
\end{tabular}
\caption{
Numerical values of the fitted parameters $\alpha$ and $\beta$
entering the Gamma form~(\ref{gamma}) of the word length distribution,
and of the resulting average and most probable word lengths
$\w n$ and $n_*$ (see~(\ref{nchar})),
for all languages shown in Figure~\ref{wldplot}.}
\label{one}
\end{center}
\end{table}

In most languages, the word length distribution is observed to be asymmetric,
with a rise steeper than its fall.
As a consequence, the average word length $\w n$ is (slightly) larger than the
most probable word length $n_*$.
Furthermore, $\w n$ and $n_*$ are large enough that they can be reasonably
evaluated by treating $n$ in~(\ref{gamma}) as a continuous variable.
This yields $C=\beta^{\alpha+1}/\Gamma(\alpha+1)$ and
\beq
\w n\approx\frac{\alpha+1}{\beta},\quad
n_*\approx\frac{\alpha}{\beta},
\label{nchar}
\eeq
where it is understood that the second estimate is rounded to the nearest integer.
The resulting numerical values of $\w n$ and $n_*$ are also given in Table~\ref{one}
for all languages shown in Figure~\ref{wldplot}.
For the European languages we have chosen, the average word length increases
consistently from English through to Finnish, as shorter words are more and
more suppressed;
this is reflected by an increase of $\alpha$, while $\beta$ varies only mildly.
On the other hand, for the Asian languages we have chosen,
both $\alpha$ and $\beta$ show appreciable variations.

For our purposes, the relevant concept of a word length is the number of its
sounds, since we are here concerned with hearings and mishearings.
Fortunately, it turns out that there is a high degree of correlation between
the length distributions
of written words (numbers of letters) and spoken words (numbers of `sounds',
which here represent phonemes), for most languages.
As a matter of fact, sounds and letters enjoy a near one-to-one correspondence
in ancient languages like Latin or Greek,
so that their length distributions for written and spoken words are virtually identical.
We would therefore expect that this high degree of correspondence between
phonology and orthography would persist in modern languages; and have, in order
to confirm this expectation,
performed a preliminary analysis of this question in the case of three
languages (German, Dutch and English),
based on the CELEX database~\cite{celex}.
Our main conclusion is that there is a fixed ratio $r\approx0.85$ between the
numbers of sounds and letters in typical words~\cite{ustocome}:
we obtain nearly identical values of $r$ for unique words (when every word of
the lexicon is counted once)
and token words (when words are counted according to their frequencies of occurrence).
Furthermore, this value of the ratio $r$ appears to have a high degree of
universality and in particular,
does not show any appreciable difference among the three languages we have tested.

We will, in what follows, therefore use the length distribution of written
words (measured by numbers of letters
and parametrised by the Gamma distribution~(\ref{gamma})) as a proxy for that
of spoken words (measured by numbers of sounds).
Our framework of the modelling of mishearings and the ensuing statistics of
word variants (derived in Section~\ref{variants}) will also be extensively used
in the next sections, where we will continue with our assumption of a
structureless lexicon.
The only language-specific ingredients entering our work will be the parameters
$\alpha$ and $\beta$ in the parametrisation~(\ref{gamma}) of the word length distribution,
which are listed in Table~\ref{one} for all languages shown in Figure~\ref{wldplot}.

\section{The statics of word recognition}
\label{wordrec}

\subsection{Decrypting words -- the easy-to-hard transition}
\label{e2h}

A key feature of speech decryption is that the number of variants induced by
mishearings should not be too large.
In this section we first define a threshold $\delta_n$,
which determines the difficulty of decryption in terms of a simple ratio
involving the numbers of words and their variants.
Another meaningful quantity which we introduce in this context is $\delta_{n_\st}$,
which describes the onset of word lengths which are so large that decryption
becomes very difficult;
this occurs both because the words themselves are exceptionally long and therefore rare,
and because they generate huge numbers of variants.
Finally, we introduce a mechanism by which most words are actually recognised:
this is the phenomenon of {\it anticipation} where, part way through a word, a
listener can guess what it actually is.
In the concluding parts of this section, we quantify the phenomenon of anticipation:
first, in the absence of mishearings, and then, in their presence.
Our results demonstrate that the phenomenon of word anticipation is much more
efficient in the former case compared to the latter,
which is intuitively reasonable.

For a language with an overall lexicon size $\Lambda$
and an arbitrary word length distribution $p_n$,
we expect that it would be easy to recognise a word $w$ of length $n$
if the average number $\Omega_n$ of its variants is much smaller than the total
number
\beq
W_n=\Lambda p_n
\eeq
of words of length $n$.
This criterion is conveniently measured by the effective exponent
\beq
\delta_n=\frac{\ln\Omega_n}{\ln W_n}=\frac{\kappa n}{\ln\Lambda p_n}.
\label{deltan}
\eeq
This quantity has a rich non-linear dependence on the word length $n$.
Its numerator grows linearly with $n$, as a consequence of~(\ref{omegaave}),
with $\kappa$ defined in~(\ref{kappa}).
Its denominator involves the full word length distribution, parametrised as~(\ref{gamma}),
with language-specific parameters $\alpha$ and $\beta$.
The introduction of the $n$-dependent scaling exponent $\delta_n$ is one of the
novel features of our work,
allowing for the application of concepts stemming from finite-size scaling.
This demarcation of complexity by size is strikingly reminiscent of problems
encountered in combinatorial optimisation~\cite{garey,shieber,monasson,kms}.

For the typical parameter values $q=0.2$ and $\Lambda=10^5$,
the effective exponent~$\delta_*$,
corresponding to the most probable word length $n_*$
of all languages shown in Figure~\ref{wldplot},
is tabulated in Table~\ref{two}.
We note that, while $\delta_*$ is small, it is significantly larger than the
corresponding value~(\ref{deltazero})
for a free lexicon with no constraint on the word length distribution;
this suggests that constraints make word identification more difficult, as
might be expected.

\begin{table}[!ht]
\begin{center}
\begin{tabular}{|l|c|c|c|}
\hline
Language & $\delta_*$ & $n_\st$ & $m_\ant$ \\
\hline
English & 0.14 & 25 & 4 \\
French & 0.16 & 26 & 5 \\
Hungarian & 0.20 & 27 & 7 \\
Finnish & 0.24 & 29 & 8 \\
\hline
Korean & 0.09 & 15 & 3 \\
Hindi & 0.11 & 20 & 4 \\
Tagalog & 0.15 & 23 & 5 \\
Burmese & 0.19 & 32 & 6 \\
\hline
\end{tabular}
\caption{
Values of characteristic quantities pertaining to our static approach to
speech recognition (see Sections~\ref{e2h} and~\ref{anti0}),
for all languages shown in Figure~\ref{wldplot}:
$\delta_*$ is the effective exponent $\delta_n$ of~(\ref{deltan}) measured
for the most probable word length $n_*$,
whereas $n_\st$ is the static crossover length such that $\delta_{n_\st}=1$
(see~(\ref{nstdef})),
and $m_\ant$ is the anticipation length in the absence of mishearings
(see~(\ref{mantdef})).
Word length distributions are parametrised according to~(\ref{gamma}),
with fitted parameters $\alpha$ and $\beta$ given in Table~\ref{one}.
Other parameter values are $q=0.2$, $\Lambda=10^5$ and $\eps=1/20$.}
\label{two}
\end{center}
\end{table}

For very long words, two effects appear.
First, the length distribution $p_n$ falls off very fast, so that the
corresponding words are genuinely rare;
second, the sheer length of the word generates hugely many variants in the
presence of mishearings.
Eq.~(\ref{deltan}) shows that these parallel effects cause
the effective exponent $\delta_n$ to increase rapidly with the word length $n$.
Word recognition becomes very hard when $n$ reaches the static crossover
length~$n_\st$, such that
\beq
\delta_{n_\st}=1.
\label{nstdef}
\eeq

This crossover length, where speech recognition becomes suddenly hard,
is analogous to the occurrence of the static easy-to-hard phase transition
met with in typical hard combinatorial
problems~\cite{garey,shieber,monasson,kms}.
Since word lengths are always finite, any potential phase transition is of course
rounded to a crossover; also, and equally clearly, its exact location depends
on the number chosen on the right-hand side of the definition~(\ref{nstdef}).
However, and despite these quantitative caveats, the analogy of the
crossover~(\ref{nstdef}) with the easy-to-hard combinatorial phase transition
is striking.

The static crossover length $n_\st$ defined in~(\ref{nstdef}) is also given in
Table~\ref{two} for all languages shown in Figure~\ref{wldplot}:
note that this is typically three times larger than the average word length $\w
n$ in each case (see Table~\ref{one}).
Words that are equal to or larger than this value in length are in the tails of
the distribution;
while they are not numerous -- typically no more than a few hundreds --
their rarity adds to the complexity of their decryption, as mentioned above.

\subsection{Word anticipation with no mishearings}
\label{anti0}

How do we decrypt individual words? A major role is played by the phenomenon of
{\it anticipation}, where we expect to hear a particular word
on the basis of the first few sounds that we hear.
In this section, we will assume that both speakers and listeners
are perfect, i.e., that no mishearings occur, while in the following one, we
will see what happens when listeners mishear what has been said.

Assume the listener has heard a string
\beq
s=\s_1\dots\s_m
\eeq
of $m$ sounds.
The string $s$ coincides with the first $m$ sounds of some set of words~$w$ of
the lexicon
of length $n\ge m$; we therefore say that it is a prefix of each of those
words.

The number $S_n$ of word prefixes of length $n$
can be estimated by merely assuming that each prefix $s$ has a small
probability $\eps$ to be an exact word, with~$\eps$ being of the order of $1/\nu$.
This yields
\beq
S_n=\frac{W_n}{\eps}=\frac{\Lambda p_n}{\eps}.
\eeq

The average number $f_m$ of words with the string $s$ as a prefix is given by:
\beq
f_m=\frac{1}{S_m}\sum_{n\ge m}W_n=\frac{\eps}{p_m}\sum_{n\ge m}p_n.
\eeq

As the number $m=1,2,3,\dots$ of heard sounds increases, the number of
remaining possibilities decreases sharply.
Usually, one is able to guess the word at a certain point before the word has
ended: we call this the {\it anticipation length}~$m_\ant$,
at which the number of remaining words becomes of the order of one:
\beq
f_{m_\ant}=1.
\label{mantdef}
\eeq
The anticipation length $m_\ant$ is also given in
Table~\ref{two} for all languages shown in Figure~\ref{wldplot}.
This length is always smaller than the most probable word length $n_*$,
as might be expected for an anticipation effect.
This concept, to which we give a quantitative underpinning here,
is referred to as the uniqueness point in linguistics,
e.g.~in cohort models~\cite{lahiri2,wdmw}.
The strength of the anticipation effect is measured by the ratio $m_\ant/n_*$.
For the choice of $\eps=1/20$, and for the languages we have considered,
this ratio is nearly constant,
suggesting that we can predict a typical word about two-thirds of the way
through it.

\subsection{Word anticipation in the presence of mishearings}
\label{anti1}

Everyday experience tells us that mishearings make it difficult to recognise,
never mind anticipate, spoken words.
We would therefore expect that with enough mishearings, the anticipation of words would
disappear entirely, which is indeed what our formalism predicts.

\begin{figure}[!ht]
\begin{center}
\includegraphics[angle=0,width=.55\linewidth,clip=true]{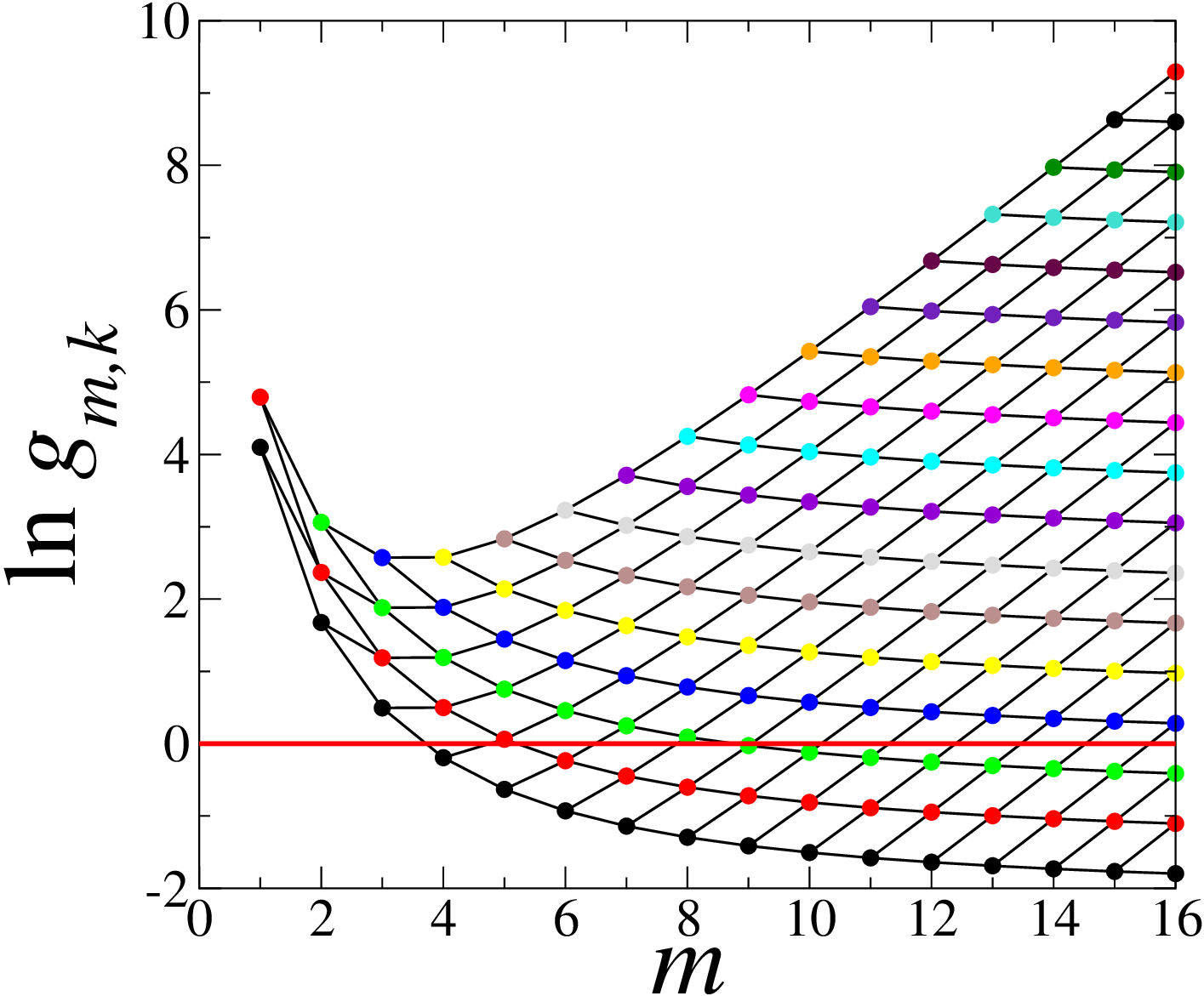}
\vskip 6pt
\includegraphics[angle=0,width=.55\linewidth,clip=true]{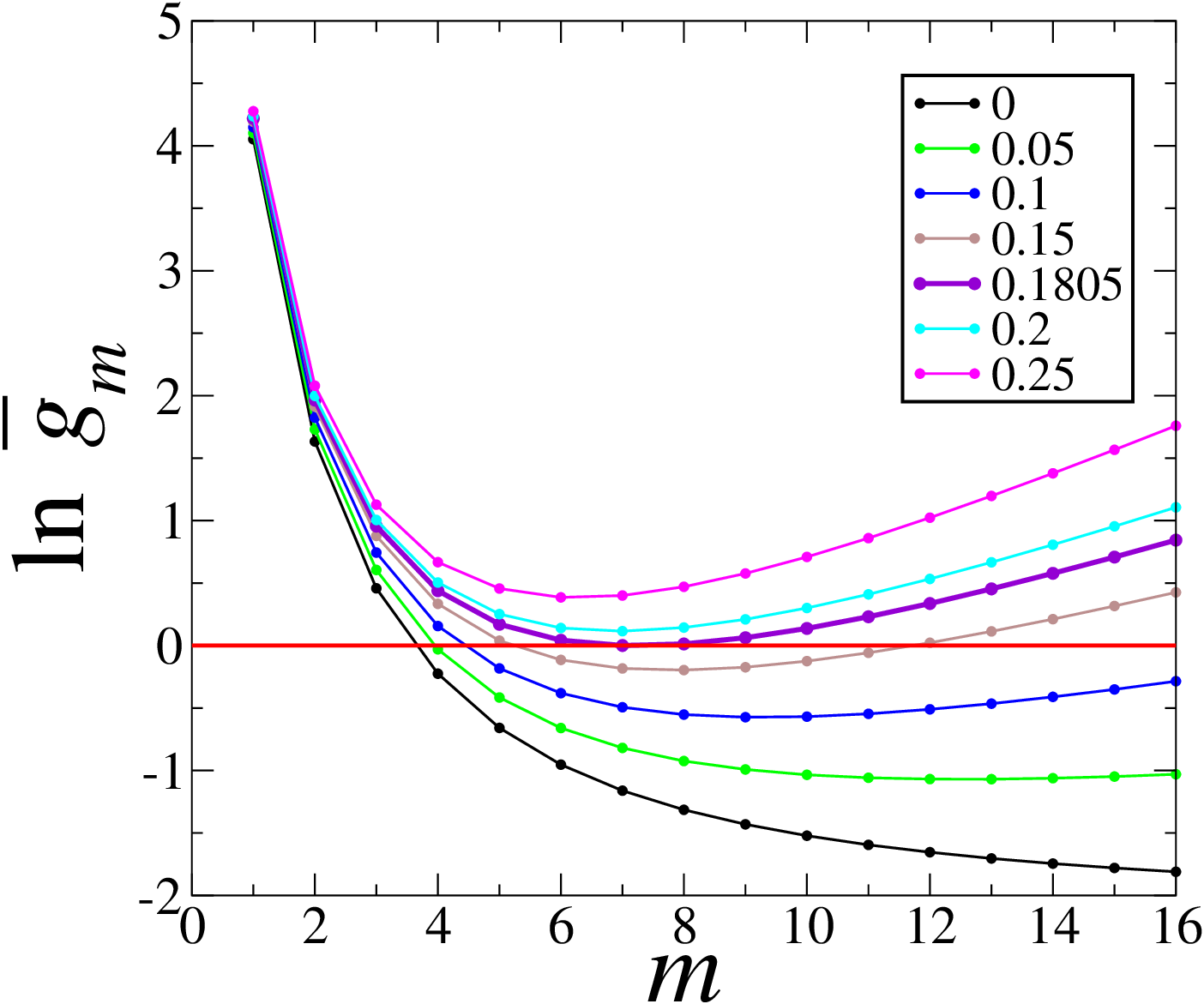}
\caption{
Plots of $\ln g_{m,k}$ (upper panel) and of $\ln\w g_m$ (lower panel) against
prefix length $m$
for model parameters corresponding to English (see~Table~\ref{one}) and
$\eps=1/20$.
In the upper panel, the number $k=0,\dots,m$ of mishearings is denoted by the
symbol colours (black, red, green, blue, etc.~denote $k=0,1,2,3,\dots$).
In the lower panel, colours denote values of the mishearing probability $q$
(see legend).
The thick purple curve corresponds to the threshold value $q_\th=0.1805$.}
\label{hammock}
\end{center}
\end{figure}

From now on, we consider the situation where each sound is misheard with
probability $q$,
according to the rules~(\ref{rules}).
Assume the listener has heard a string
\beq
t=\t_1\dots\t_m
\eeq
of $m$ sounds.
This string may be the outcome of the random insertion of any number
$k=0,\dots,m$ of mishearings into an existing word prefix $s$ of length $m$.
For a fixed number $k$ of mishearings,
the mean number of words $w$ compatible with the heard prefix $t$ reads therefore
\beq
g_{m,k}=2^kf_m.
\label{gmkres}
\eeq
This expression can be visualised as a `hammock plot' (upper panel of Figure~\ref{hammock})
of~$g_{m,k}$ against the prefix length $m$, for model parameters corresponding
to English (see~Table~\ref{one}) and $\eps=1/20$.
The number $k$ of mishearings is denoted by the symbol colours.
The number of words varies over so large a range that a logarithmic scale is needed.

As before, the anticipation threshold is defined as the first point at which
only one word remains possible.
This amounts, in the present instance, to setting $g_{m,k}=1$,
which is shown in the upper panel of Figure~\ref{hammock} as a thick red
horizontal line.
Note that for the case of no mishearings, we have $k=0$ and $g_{m,k}=f_m$,
corresponding to the lowest branch of the plot (black symbols).
The fact that this crosses the red line at $m=4$ provides a consistency check
with the fact that the onset of the
anticipation effect for English in the absence of mishearings, $m_\ant$, is
indeed~4 (see~Table~\ref{two}).

The average number $\w g_m$ of remaining possible words after a sequence $t$ of
length~$m$ is heard can be evaluated by averaging the expression~(\ref{gmkres})
over the known distribution of the number $k$ of mishearings (see~(\ref{binom})):
\beq
\w g_m=\e^{\kappa m}f_m.
\eeq
As the mishearing probability $q$ increases,
the exponential factor $\e^{\kappa m}$ in the above expression increases rapidly,
leading to a progressive weakening of the anticipation effect
until it disappears above some threshold $q_\th$.
This threshold is defined as the point at which the minimum of $\w g_m$ over
all prefix lengths $m$ becomes unity.
In the plot of $\ln\w g_m$ against the prefix length $m$ in the lower panel of
Figure~\ref{hammock},
this happens for the third curve from the top (thick purple curve),
where the minimum grazes the red line corresponding to $\w g_m=1$.
The corresponding threshold reads $q_\th\approx0.18$:
for $q$ less than this value (lower four curves of the plot), $\w g_m$ always
crosses the red line,
so that anticipation is possible even in the presence of mishearings.
On the other hand, for $q$ larger than $q_\th$ (upper two curves of the plot),
$\w g_m$ stays above 1,
so that there are too many variants for anticipation to be effective.

Table~\ref{three} gives the numerical values of the threshold mishearing probability
$q_\th$ for all languages shown in Figure~\ref{wldplot},
as well as the corresponding values of the prefix length $m_\th$ where the minimum
is reached at threshold and of the effective exponent~$\delta_\th$.
For all languages considered here $m_\th$ and $n_*$ are nearly equal
(compare Table~\ref{one} and Table~\ref{three}),
suggesting that the most probable word length $n_*$ in a given language
nearly optimises the possibility of decryption despite mishearings.
The corresponding effective exponent~$\delta_\th$ is much less than unity,
the value of $\delta$ at the static easy-to-hard transition~$\delta_{n_\st}$.
This important distinction makes it clear that while $\delta_{n_\st}$
epitomises the difficulty of recognising very long and complex words,
$\delta_\th$ corresponds to the befuddling effect of multiple mishearings,
which can set in for words of even moderate sizes.

\begin{table}[!ht]
\begin{center}
\begin{tabular}{|l|c|c|c|}
\hline
Language & $q_\th$ & $m_\th$ & $\delta_\th$ \\
\hline
English & 0.18 & 7 & 0.12 \\
French & 0.15 & 9 & 0.14 \\
Hungarian & 0.12 & 12 & 0.14 \\
Finnish & 0.09 & 14 & 0.14 \\
\hline
Korean & 0.45 & 5 & 0.19 \\
Hindi & 0.26 & 6 & 0.15 \\
Tagalog & 0.18 & 8 & 0.14 \\
Burmese & 0.08 & 11 & 0.09 \\
\hline
\end{tabular}
\caption{
Values of characteristic quantities pertaining to our approach to
word anticipation in the presence of mishearings (see Section~\ref{anti1}),
for all languages shown in Figure~\ref{wldplot}:
$q_\th$ is the threshold value of the mishearing probability at which the
anticipation effect disappears,
whereas $m_\th$ and $\delta_\th$ are respectively the corresponding prefix
length and effective exponent.
Word length distributions are parametrised according to~(\ref{gamma}),
with fitted parameters $\alpha$ and $\beta$ given in Table~\ref{one}.
Other parameter values are $\Lambda=10^5$ and $\eps=1/20$.}
\label{three}
\end{center}
\end{table}

\section{The dynamics of word recognition}
\label{dynamic}

In the above, we have only considered static aspects of word recognition.
Here we turn our attention to the dynamics of this process:
more specifically, what is the effect of mishearings
on the time it takes for a listener to recognise a given word?
Does it matter if these mishearings appear individually or if they are consecutive,
and form a cluster of mishearings?
Recalling that until now we have only addressed the total number of mishearings
in a given word,
with no positional information, the following intuitive approach at least gives
us a hint of the difference between individual and collective impediments to word
recognition.

We proceed in the spirit of an analogy with jammed granular media
(see e.g.~\cite{pnasus} and references therein),
an example of an athermal and disordered system like our own.
Prior to jamming in a shaken assembly of grains,
a tracer grain can navigate its way past individual grains that only partially
obstruct it;
however, as the density increases, the obstacles represented by clusters of
grains make it impossible for it to move beyond them,
and this complete obstruction leads to jamming.
In the present scenario, a low density of mishearings, whether isolated or contiguous,
leads to a reasonable possibility of `escape' for a listener from a situation
of complete misunderstanding.
In contrast, when there are many clusters of consecutive mishearings,
the large number of variants essentially makes decryption impossible.

More specifically, we visualise the speech recognition algorithm as a random
walker on a network with traps,
where the latter correspond to mishearings.
A low density of traps causes the slowing down of the random walker;
as the density increases, the traps are more and more likely to occur
contiguously, in clusters.
In the latter case, dynamical arrest may result if the random walker has to
traverse paths crossing arbitrarily large clusters of traps.

To reiterate -- we have so far characterised speech recognition from a static viewpoint,
focusing on the statistics of mishearings in a word $w$ of length $n$.
This was done by computing the average number~$\Omega_n$ of variants per word of length $n$
(see~(\ref{omegaave})),
and the corresponding static exponents $\delta$ (see~(\ref{delta})) or
$\delta_n$ (see~(\ref{deltan})).

We now approach the problem from a dynamical point of view: how does a
speech recognition algorithm react to a string of sounds that are being
successively presented to it in real time? The analogies given above suggest
that a cluster of successive mishearings would clearly present a far greater
obstacle to speech recognition than a few isolated ones.
We therefore proceed by attributing a (dynamical) penalty, in the form of a
statistical weight $\lambda_k$, to any set of $k$ successive mishearings.
This penalty should increase rapidly with the cluster size~$k$, as larger
clusters of mishearings will more significantly slow down the task of decryption.
It should also be proportional to the complexity associated with the task of
exploring such a cluster in an ordered way, and thus to the number of its
possible permutations.

This intuitive line of reasoning leads us to associate cluster weights that
grow factorially with a cluster of $k$ consecutive mishearings:
\beq
\lambda_k=\mu^k\,k!.
\label{lamfact}
\eeq
The constant $\mu$ is a phenomenological parameter which cannot be estimated {\it a priori}.

The full dynamical complexity of deciphering a typical word of length $n$ is
thus represented by the partition function~$Z_n$
associated with a chain of length $n$ in the presence of the cluster weights $\lambda_k$.
From the formal definition of these quantities
and their subsequent analysis in~\ref{append}, we obtain a logarithmic
growth of the free energy density (see~(\ref{knlapp})):
\beq
K_n\approx\ln\frac{\mu q n}{\e}.
\eeq
The resulting superextensive growth of the total free energy $\ln Z_n=nK_n$
implies that the effective dynamical exponent,
\beq
\Delta_n=\frac{\ln Z_n}{\ln W_n}=\frac{nK_n}{\ln\,\Lambda p_n},
\label{Deltan}
\eeq
grows more quickly with word length $n$ than its static counterpart $\delta_n$
(see~(\ref{deltan})).
Proceeding to define the dynamical crossover length $n_\dy$ by the condition
\beq
\Delta_{n_\dy}=1,
\label{ndydef}
\eeq
we assert that for large enough word lengths and sufficiently high $q$,
large clusters of mishearings will arrest the speech recognition algorithm at $n_\dy$,
well before the static crossover length $n_\st$ is reached (see below).

The above ideas have a wealth of analogies in statistical and condensed matter physics.
In addition to the dynamical arrest in granular media referred to above,
similar dynamical transitions in diffusive motion have been widely observed in glasses
(see e.g.~\cite{kirkpatrick,marinari,castellani,krzakala}),
as well as via the occurrence of Anderson localisation,
be this in disordered conductors (see e.g.~\cite{KM,fifty}) or in the multiple
scattering of light (see e.g.~\cite{vrn,segev}).
From a more theoretical point of view, a well-documented instance of slowing
down from diffusive to sub-diffusive transport
is provided by random walks on comb structures, in the regime where the mean
value of the teeth depth diverges (see e.g.~\cite{havlin}).

In summary, our intuitive arguments above suggest that decryption in the
presence of mishearings
involves two successive thresholds as a function of word length~$n$,
viz.~the dynamical length $n_\dy$~(\ref{ndydef}) and the static one $n_\st$~(\ref{nstdef}).
Both crossover lengths depend on the lexicon size $\Lambda$, the mishearing probability $q$,
and the language under consideration, through the parametrisation~(\ref{gamma})
of the word length distribution.
A subtlety is that the dynamical crossover length $n_\dy$ also depends on the
unknown parameter $\mu$.
The upper limit of this dynamical length is naturally provided by the static length $n_\st$,
to which it becomes equal at a critical value~$\mu_c$.
Whenever $\mu>\mu_c$, the dynamical length $n_\dy$ exhibits a slow decrease as
a function of $\mu$;
at any particular value of the latter, the dynamical transition precedes the
static transition, as it must.
In the opposite regime ($\mu<\mu_c$), both static and dynamical transitions
merge into a single one.

\begin{figure}[!ht]
\begin{center}
\includegraphics[angle=0,width=.55\linewidth,clip=true]{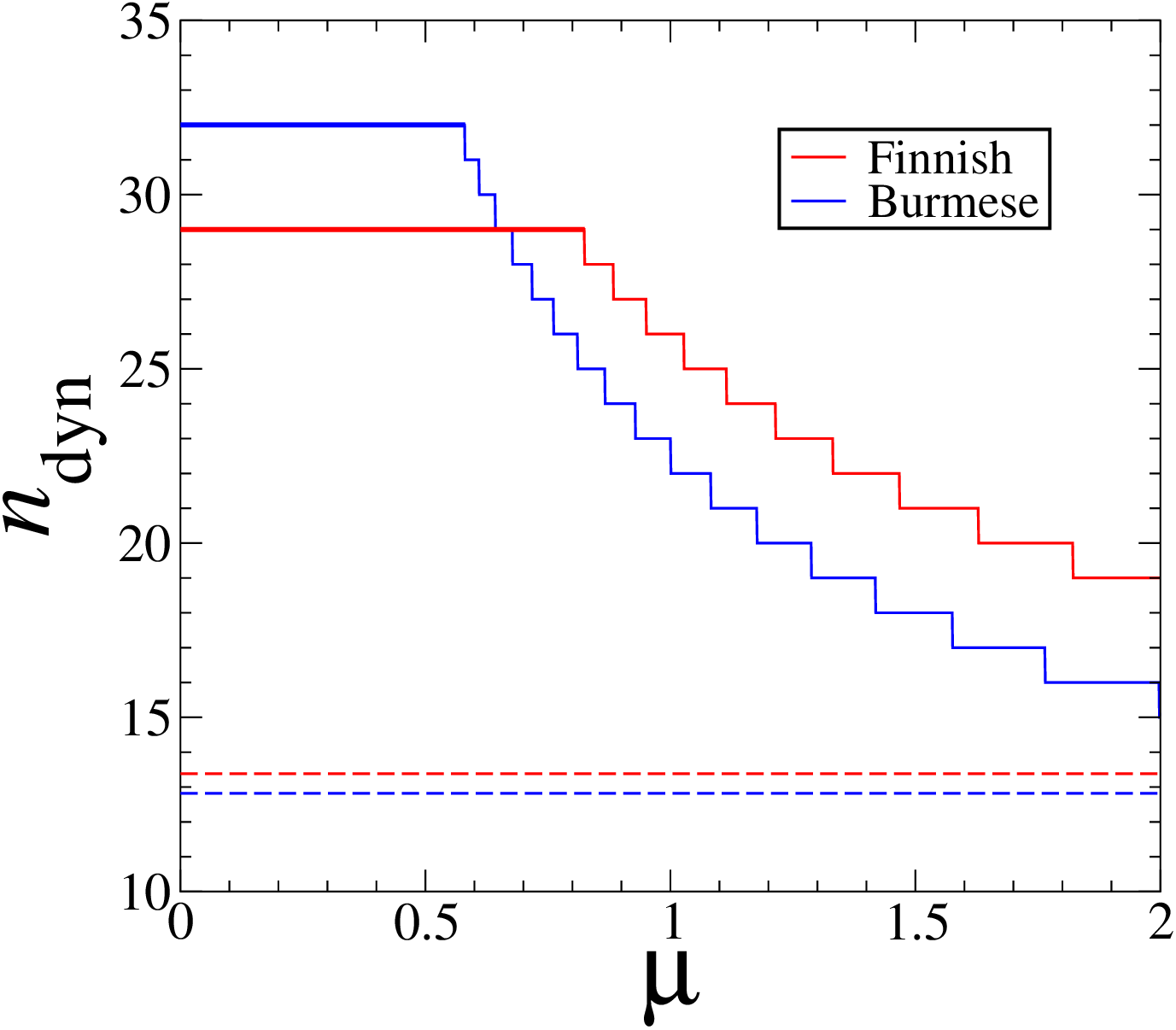}
\caption{
Dynamical crossover length $n_\dy$ against parameter~$\mu$,
for Finnish and Burmese.
Thick full horizontal lines: static crossover length~$n_\st$.
Dashed horizontal lines: average word length $\w n$.
Word length distributions are parametrised according to~(\ref{gamma}),
with fitted parameters $\alpha$ and $\beta$ given in Table~\ref{one}.
Other parameter values are $q=0.2$ and $\Lambda=10^5$.}
\label{statdyn}
\end{center}
\end{figure}

This is demonstrated in Figure~\ref{statdyn} for two languages, Finnish and Burmese,
which we have chosen because they have relatively high values of $n_\st$
(see~Table~\ref{one}).
The dynamical length $n_\dy$ varies over a broad range extending from
the average word length $\w n$ to the static crossover length $n_\st$,
where the latter corresponds to word lengths deep in the tails of the distribution.
For languages such as English, the picture is qualitatively similar,
although the range over which $n_\dy$ varies is smaller.

\section{Discussion}
\label{discussion}

We have in the above built a simple and intuitive model of speech perception.
A unique feature of our model is the representation of mishearings via a single parameter,
the mishearing probability $q$, providing a measure of the fraction of sounds
that can be misheard in a given language.
We go on to obtain intuitive predictions in a series of scenarios relating to
word recognition.
In the absence of mishearings, we predict the existence of a threshold after
which a word can be correctly guessed, viz.~the anticipation length~$m_\ant$;
this provides a quantitative underpinning to a related concept in linguistics,
the uniqueness point.
We follow this with the consideration of anticipation in the presence of mishearings;
as expected the phenomenon disappears above a threshold mishearing probability~$q_\th$.

A limitation of our current approach is that it proceeds by considering the
fate of `typical' words,
thereby following a mean-field approach, where the role of fluctuations as well
as correlations is neglected.
Thus, we are so far unable to include positional information in our analysis --
e.g.~the relative difficulty
of decryption depending on whether mishearings occur at the beginning or the
end of a word~\cite{lahiri3}, which is clearly important for real-world decryption.
We aim to include the effect of such and other correlations in future work~\cite{ustocome}.

A leitmotif underpinning our work is the idea that speech perception is a
problem of optimisation.
We have identified a {\it static easy-to-hard transition} in terms of word length,
after which spoken words are very hard to decipher.
This occurs for words in the tails of the word length distribution, whose
lengths are greater than the threshold static length $n_\st$; we have defined
$\delta_{n_\st}$ as a measure
of the difficulty of decryption in this limit of long and typically rare words.
In the presence of mishearings, however, decryption can be difficult for words
of even moderate size; the quantity $\delta_\th$ defined above embodies
the difficulty of decryption in the presence of multiple mishearings.

The dynamics of speech recognition takes these ideas a step further, and
examines the relative difficulties
presented by isolated versus consecutive mishearings to word recognition:
everyday experience
tells us that clusters of consecutive mishearings are likely to provide far
greater impediments to
decryption than isolated ones.
The dynamical complexity of such clusters increases with their size,
a feature which is modelled by the association of appropriate statistical
weights with them.
The task of the speech recognition algorithm is to navigate these obstacles and
correctly identify the word concerned,
which becomes unfeasible beyond some dynamical length~$n_\dy$,
where clusters of contiguous mishearings can successfully arrest recognition.
This phenomenon is associated with a {\it dynamical transition} which, as in the
case of many other complex systems, precedes the static transition occurring at $n_\st$.

We suggest that the dynamical transition is intimately associated with the phenomenon
of `underspecification' proposed by Lahiri et al~\cite{lahiri1}.
Put simply, this involves the storage
of all possible word variants corresponding to a set of mishearings in the
listener's memory, until
an individual word is correctly recognised at a suitable point in a sequence of sounds.
The presence of clusters of contiguous mishearings will cause this list to be hugely
amplified, until accurate word recognition is impossible.
Clearly, this can happen even for words close to the average word length $\w n$, i.e.,
the dynamical transition~$n_\dy$ can set in for word lengths that vary between $\w n$
and the static crossover length $n_\st$.

A possible experiment to test this would involve tests carried out on ensembles
of listeners with similar linguistic abilities in a given language,
who would be subjected to ever-increasing clusters of contiguous mishearings
in individual words; datasets would comprise words with lengths between $\w n$ and $n_\st$.
The objective would be to identify a threshold of incomprehension for most listeners
which, if found, would represent the dynamical transition~$n_\dy$.
Our work in this paper suggests that this should set in
well before the static crossover length $n_\st$, for typical world languages.

\section*{Acknowledgments}

We are grateful to Peter Stadler for having given us access to the Leipzig
Corpora Collection,
and to Aditi Lahiri and Henning Reetz for having made us aware of the CELEX database.
AM warmly thanks the Leverhulme Trust for the
Visiting Professorship that funded this research, as well as the Faculty
of Linguistics, Philology and Phonetics at the University of Oxford, for their hospitality.

\appendix

\section{Chains with intra-cluster interactions}
\label{append}

This appendix is devoted to the evaluation of the partition function $Z_n$ of
a binary chain of length $n$ with arbitrary intra-cluster interactions.
This quantity enters the definition~(\ref{Deltan}) of the effective dynamical
exponent $\Delta_n$.

We consider the setting of the site percolation problem (see e.g.~\cite{SA}).
Most notations are chosen for the sake of consistency with the body of the paper.
Each site is either occupied with probability $q$,
or empty with the complementary probability $p=1-q$.
Furthermore, arbitrary statistical weights $\lambda_k$ are attached to each cluster
of $k\ge1$ consecutive occupied sites:
\beq
{\circ\atop\null}\,
{\underline{\bullet\;\bullet}\atop^{\lambda_2}}\,
{\circ\,\circ\,\circ\atop\null}\,
{\underline{\bullet}\atop^{\lambda_1}}\,
{\circ\;\circ\atop\null}\,
{\underline{\bullet\,\bullet\,\bullet}\atop^{\lambda_3}}\,
{\circ\atop\null}\,
{\underline{\bullet\;\bullet}\atop^{\lambda_2}}\,
{\circ\;\circ\atop\null}\,
{\underline{\bullet\,\bullet\,\bullet\;\bullet}\atop^{\lambda_4}}\,
{\circ\;\circ\atop\null}\,
\eeq

The total weight of a configuration is the product of the weights $\lambda_k$
over all clusters.
The partition function $Z_n$
is the sum of these total weights over the~$2^n$ configurations
of an open finite chain of $n$ sites,
where each configuration is attributed a probability stemming from the site
percolation problem.
The partition function can be evaluated by means of recursion relations,
somewhat along the lines of the transfer-matrix formalism (see e.g.~\cite{baxter,CPV}).
In the present situation, it is useful to write
\beq
Z_n=Z_n^\circ+Z_n^\bullet,
\eeq
where $Z_n^\circ$ (resp.~$Z_n^\bullet$)
are the partial partition sums
over configurations whose leftmost site is empty (resp.~occupied).
The latter quantities obey the recursion relations
\beq
Z_n^\circ=p^n+\sum_{m=1}^{n-1}p^mZ_{n-m}^\bullet,\quad
Z_n^\bullet=\lambda_nq^n+\sum_{m=1}^{n-1}\lambda_mq^mZ_{n-m}^\circ.
\eeq
The above equations are self-explanatory.
The summation index is nothing but the size $m=1,\dots,n-1$ of the leftmost cluster
of empty sites in the first equation,
of occupied sites in the second one.
These recursion equations can be solved by introducing the generating series
\beq
G^\circ(z)=\sum_{n\ge1}Z_n^\circ z^n,\quad
G^\bullet(z)=\sum_{n\ge1}Z_n^\bullet z^n,
\eeq
which obey
\beq
G^\circ(z)=\frac{pz}{1-pz}(1+G^\bullet(z)),\quad
G^\bullet(z)=L(z)(1+G^\circ(z)),
\eeq
with
\beq
L(z)=\sum_{k\ge1}\lambda_kq^kz^k.
\eeq
We thus obtain
\beq
G^\circ(z)=\frac{pz(1+L(z))}{1-pz(1+L(z))},\quad
G^\bullet(z)=\frac{L(z)}{1-pz(1+L(z))},
\eeq
and finally
\beq
G(z)=\sum_{n\ge1}Z_nz^n=G^\circ(z)+G^\bullet(z)
=\frac{pz+(1+pz)L(z)}{1-pz(1+L(z))}.
\label{gres}
\eeq

We are mostly interested in the asymptotic growth of the partition function~$Z_n$
for large chain lengths $n$.
This growth law obeys the following dichotomy:

\begin{itemize}

\item[$\bullet$]
If the cluster weights $\lambda_k$ are bounded by an exponential of $k$,
the generating series $L(z)$ has a non-zero radius of convergence.
As a consequence, the partition function~$Z_n$ grows exponentially with the
chain size~$n$.
This is in agreement with the existence of a well-defined thermodynamic limit
where the total free energy is extensive,~i.e.,
\beq
\ln Z_n\approx nK.
\eeq
The free energy density $K$ per site is given
by the property that
\beq
z_*=\e^{-K}
\eeq
is the smallest zero of the denominator of~(\ref{gres}).

It is worth considering a few examples.

For exponential cluster weights $\lambda_k=\mu^k$,
introducing these weights boils down to the renormalisation of $q$ into
the product $\mu q$.
We accordingly obtain the simple result
\beq
K=\ln(p+\mu q).
\eeq

For linear cluster weights $\lambda_k=ak$,
we obtain a cubic equation for $z_*$:
\beq
(1-pz_*)(1-qz_*)^2=apqz_*^2.
\eeq

For quadratic cluster weights $\lambda_k=bk^2$,
we obtain a quartic equation:
\beq
(1-pz_*)(1-qz_*)^3=bpqz_*^2(1+qz_*).
\eeq
The two latter examples
demonstrate that the expression of the free energy soon becomes quite intricate.

\item[$\bullet$]
If the cluster weights $\lambda_k$ grow faster than any exponential,
the generating series $L(z)$ has a vanishing radius of convergence.
In such a circumstance,
the partition function is dominated by the configuration where all sites are occupied,
i.e.,
\beq
Z_n\approx q^n\lambda_n.
\label{znres}
\eeq
The first correction to this leading result originates in the two configurations
where only the first (or last) site is empty, yielding
\beq
Z_n=q^n\lambda_n\left(1+\frac{2p\lambda_{n-1}}{q\lambda_n}+\cdots\right).
\eeq
The correction term is negligible for large chain sizes,
precisely because the cluster weights have a superexponential growth.
The result~(\ref{znres}) can be recast as
\beq
\ln Z_n=nK_n,
\eeq
where the effective free energy is superextensive, as its density
\beq
K_n\approx\ln q+\frac{\ln\lambda_n}{n}
\eeq
keeps growing indefinitely with the chain size~$n$.

\end{itemize}

\begin{figure}[!ht]
\begin{center}
\includegraphics[angle=0,width=.55\linewidth,clip=true]{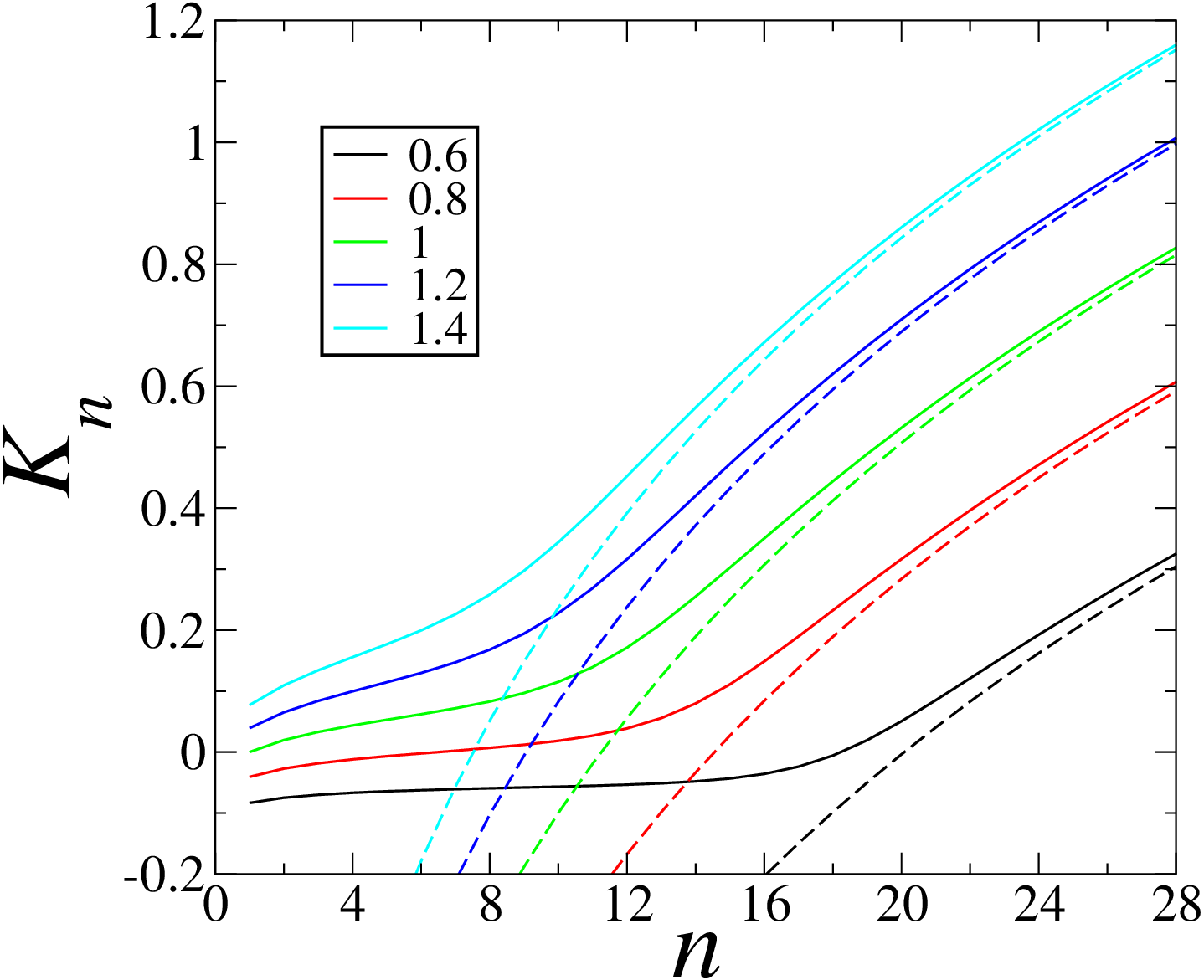}
\caption{
Free energy density $K_n$ against chain size $n$,
for $q=0.2$ and factorially growing cluster weights (see~(\ref{lamapp})),
with several values of the parameter $\mu$ (see legend).
Dashed curves: logarithmic asymptotic growth law~(\ref{knlapp}).}
\label{kn}
\end{center}
\end{figure}

The situation where cluster weights grow factorially, according to
\beq
\lambda_k=\mu^k\,k!,
\label{lamapp}
\eeq
plays a key role in the body of this paper (see~(\ref{lamfact})).
There, the asymptotic growth of the above cluster weights is only marginally
superexponential,
as the corresponding free energy density exhibits a logarithmic growth, according to
\beq
K_n\approx\ln\frac{\mu q n}{\e}.
\label{knlapp}
\eeq
This is illustrated in Figure~\ref{kn},
showing plots of the free energy density $K_n$ against~$n$
for several values of the parameter $\mu$.

\bibliographystyle{ws-acs}
\bibliography{speech}

\end{document}